\documentclass[sigconf]{acmart}
 \pdfoutput=1

\AtBeginDocument{%
  \providecommand\BibTeX{{%
    \normalfont B\kern-0.5em{\scshape i\kern-0.25em b}\kern-0.8em\TeX}}}

\usepackage{times}
\usepackage{multirow}
\usepackage{amsmath}
\usepackage{helvet}
\usepackage{courier}
\usepackage{booktabs}
\usepackage{latexsym}
\usepackage{algorithm}
\usepackage{algorithmic}
\usepackage{graphicx}
\def\UrlFont{\rm}
\usepackage{tabularx}

\usepackage{enumerate}
\usepackage{enumitem}
\usepackage{lipsum} 

\usepackage{amsmath,amssymb,bbm,mathtools,bbold,adjustbox}
\usepackage{url}
\usepackage{bm}
\usepackage{xcolor}
\usepackage{color}
\usepackage{xspace}
\usepackage{amsthm}

\theoremstyle{definition}

\newcommand{\mname}{\texttt{DeepEnroll}\xspace}

\begin{document}

\title{\mname: Patient-Trial Matching with Deep Embedding and Entailment Prediction}





\author{Xingyao Zhang$^1$,  Cao Xiao$^2$, Lucas M. Glass$^{2,3}$,  Jimeng Sun$^{4,5}$}

\email{
 xingyao-17@mails.tsinghua.edu.cn, {cao.xiao,lucas.glass}@iqvia.com,jimeng@illinois.edu
}
\affiliation{Tsinghua University$^1$,  IQVIA$^2$, Temple University$^3$,  Georgia Institute of Technology$^4$,\\ University of Illinois Urbana-Champaign$^5$}

\begin{abstract}
Clinical trials are essential for drug development but often suffer from expensive, inaccurate and insufficient patient recruitment.
The core problem of {\it patient-trial matching} is to find qualified patients for a trial, where patient information is stored in electronic health records (EHR) while trial eligibility criteria (EC) are described in text documents available on the web. How to represent longitudinal patient EHR? How to extract complex logical rules from EC? Most existing works rely on manual rule-based extraction, which is time consuming and inflexible for complex inference.
To address these challenges, we proposed \mname, a cross-modal inference learning model to jointly encode enrollment criteria ({\it text}) and patients records ({\it tabular data}) into a shared latent space for matching inference. \mname applies a pre-trained Bidirectional Encoder Representations from Transformers(BERT) model to encode clinical trial information into sentence embedding. And uses a hierarchical embedding model to represent patient longitudinal EHR. In addition, \mname is augmented by a numerical information embedding and entailment module  to reason over numerical information in both EC and EHR. These encoders are trained jointly to optimize patient-trial matching score. We evaluated \mname  on the trial-patient matching task with demonstrated on  real world datasets. \mname outperformed the best baseline by up to 12.4\% in average F1.
\end{abstract}




\maketitle

\section{Introduction}

Clinical trial enrollment is a long-standing problem for drug development.
Often insufficient patients enroll in trials despite that many of them having the target condition. The first barrier to trial participation is simple: many patients are unaware clinical trials are open and relevant for them. Luckily, the availability of  trial eligibility criteria (EC) online and the rich  collection of patient electronic health records (EHRs) data in hospitals bring a new promise to data driven automated trial enrollment.

Over the years, both rule-based systems and machine learning approaches were proposed for patient-trial matching. The rule-based systems~\cite{weng2011elixr,kang2017eliie} highly rely on laborious rule-setting and human annotations. Also they yield poor recall due to the existence of morphological variants and inadequate rule coverage. The machine learning based models perform rule extraction automatically. For example, Alicante {\it et al.,}~\cite{alicante2016unsupervised} applied unsupervised clustering methods for eligible rule extraction. Bustos {\it et al.,}~\cite{bustos2018learning} evaluated several classifiers including convolutional neural networks for EC classification in cancer domain. Criteria2Query~\cite{yuan2019criteria2query} utilized rules extracted from ~\cite{kang2017eliie} to generate matching patient definition.
Despite their improvement over rule-based systems, existing patient-trial matching methods still face the following challenges.


\begin{enumerate}[leftmargin=*]

    \item \textbf{Heterogeneous data from EC and patient EHR}. Existing methods take two steps to extract rules from EC first, which could yield  criteria that are too strict to find enough patients from EHR data.
    However, simultaneous matching of EC and EHR is difficult. ECs use unstructured natural language to describe eligibility criteria. EHR, on the other hand, use structured clinical codes to represent hospital visits of patients. There is a lack of cross-modal representation that can link matched concepts in ECs and EHR.
    \item \textbf{Lack of explicit modeling for numerical information}.
    EC statements often include many numerical information such as age, values of lab results and medication dosage.  It creates challenges for matching since numbers are a common source of contradictions in natural language processing tasks~\cite{dagan2013recognizing}, and are also  insensitive in distributed representations~\cite{roy2015reasoning}.
    Existing patient-trial matching works  pay little attention to this issue.
\end{enumerate}

To fill the gap, we proposed \mname to perform patient-trial matching based on clinical trial EC and patient health data.  \mname is enabled by the following technical contributions.

\begin{enumerate}[leftmargin=*]
    \item  \textbf{Joint embedding and entailment prediction to match heterogeneous data}. \mname applies a Bidirectional Encoder Representations from Transformers (BERT~\cite{devlin2018bert}) model to encode clinical trial information. It also performs hierarchical embedding to encode patient health data. And the patient-trial matching are casted as an entailment prediction problem, where we model patient embedding as \textit{hypothesis} and trial embedding as \textit{premise}, and the objective is to predict whether a particular patient can be inferred from a given trial embedding. The  encoders for EC and EHR  and the entailment task are jointly optimized to find good matches between trials and patients.
    \item  \textbf{Numerical information entailment module to explicitly match numerical information}.
    \mname is augmented by a numerical information embedding and entailment module to explicitly reason over numbers. It proposes a numerical representation to specifically encode numbers in textual input (EC). Then a pattern-based comparison algorithm is developed for (1) extracting numbers from free-text ECs and (2) inference between extracted numerical information in EC and EHR. Then the comparison results will be used to update the matching results.
\end{enumerate}
 We evaluated \mname on both real world  clinical trial dataset and a synthetic data. We evaluated the patient-trial matching via predicting patient enrollment for trials. \mname outperformed the best state-of-the-art baselines by up to 12.4\% in average F1 and 6.8\% in PR-AUC.

\section{Related Work}

\noindent\textbf{Patient-Trial Matching} includes both rule-based and machine learning based models. Among rule-based models, EliXR~\cite{weng2011elixr} matches Unified Medical Language System (UMLS) concepts and relations via pre-defined dictionary and regular expressions. Another system called EliIE~\cite{kang2017eliie} parses and formalizes free-text EC with OMOP standard~\footnote{https://www.ohdsi.org/data-standardization/the-common-data-model/}. The rule-based models learn to structuralize named entities and relations for clinical trials only, rather than for patient-trial matching. In addition, they highly rely on rule-setting and human annotations. Also they yield poor recall due to the existence of morphological variants of ECs and inadequate rule coverage. As for machine learning based algorithms, ~\cite{alicante2016unsupervised} that applies unsupervised clustering methods for eligible rule extraction. Also ~\cite{bustos2018learning} compared CNN, SVM, and kNN as classifiers for EC classification in cancer domain. More recently, Criteria2Query~\cite{yuan2019criteria2query} is a hybrid information extraction pipeline that combines machine learning and rule-based methods to form patient criteria. Compared to the existing works, \mname is an end-to-end deep learning model that achieves significantly better performance as shown in the experiment. \\

\noindent\textbf{Pre-training Techniques.}
The goal of pre-training techniques is to provide model training with good initializations.  Recently,  language model pre-training techniques such as  \cite{peters2018deep,radford2018improving,devlin2018bert} have shown to largely improve the performance on multiple natural language processing tasks. As the most widely used one, BERT~\cite{devlin2018bert} builds on the Transformer~\cite{vaswani2017attention} architecture and improves the pre-training using a masked language model for bidirectional representation.
Previously, BERT has been applied on EHR data to leverage unlabeled data for medication recommendation~\cite{ijcai2019-825}.
In this paper, we adapt the framework of BERT and pre-train our clinical trial EC embedding using large medical corpora.\\

\noindent\textbf{Cross-modal Matching} is enabled by joint embedding learning and pairwise similarity learning. Joint embedding learning aims to find a joint latent space where  embedding of similar text and tabular data are close~\cite{li2017identity,wang2016learning}. Pairwise similarity learning focuses on similarity computation for the matching~\cite{huang2017instance,wang2018learning}. For patient-trial matching task, one EC statement could entail a group of EHR and vice versa. Therefore, similarity learning is not appropriate due to the need for capturing the many-to-many relationship.

\section{Method}

\subsection{Problem Formulation}

Patient EHR data for patient $n$ can be represented as a sequence of hospital visits $\bm{H}^{(n)}=[\bm{h}^{(n)}_{1}\cdots \bm{h}^{(n)}_{N}]$. Each hospital visit $\bm{h}^{(n)}_{i}$ consists of a major diagnosis code $\bm{d}^{(n)}_{i} \in \bm{V}_d$ and a set of associated treatments(medications or procedures) $\bm{M}^{(n)}$.Similarly, each $\bm{M}_i^{(n)}$ consists of varying number of treatments $ \bm{m}^{(n)}_{i,1},\cdots,\bm{m}^{(n)}_{i,|\bm{M}_i^{(n)}|}$.
The corresponding clinical trial $m$ can be represented as $\bm{E}^{(m)} = (\bm{e}^{(m)}_1, \cdots \bm{e}^{(m)}_M)$ where each $\bm{e}^{(m)}_i$ indicates the $i$-th EC and can be represented as a sequence of words: $\bm{e}^{(m)}_i=[\bm{a}^{(m)}_{i,1}, \bm{a}^{(m)}_{i,2},\ldots ]$ where $\bm{a}^{(m)}_{i,j}$ is the $j$-th word in EC $\bm{e}^{(m)}_i$. In this paper, we embed EC and EHR into a shared latent space, where $\bm{U} \in \mathbb{R}^{M\times o}$  and $\bm{V} \in \mathbb{R}^{N\times o}$ are the sets of distributed representation of $\{\bm{e}^{(m)}_1,\cdots \bm{e}^{(m)}_M\}$ and $\{\bm{h}^{(n)_1},\cdots \bm{h}^{(n)}_N\}$ after transformation. The matching is casted as a multiclass classification problem, where the matching results between $\{\bm{e}^{(m)}_1,\cdots \bm{e}^{(m)}_M\}$ and  $\{\bm{h}^{(n)_1},\cdots \bm{h}^{(n)}_N\}$ are classified into the following categories: "entailment", "contradiction" and "neutral".

\begin{table}[h!]
\centering
\caption{Notations used in this paper.
}\label{tb:notation}
\resizebox{1\columnwidth}{!}{
\begin{tabular}{c|l}
 \toprule
\textbf{Notation} & \textbf{Description} \\
 \midrule
 $\bm{E}^{(m)}$; $\bm{V}^{(n)}$ &  EC set of the $m$-th trial; EHR set of the $n$-th patient \\
 $\bm{e}^{(n)}_i$ & the $i$-th EC of the $n$-th trial\\
$\bm{W}_d$ &  unnormalized attention weight matrix in alignment module\\
$\bm{U} \in \mathbb{R}^{M\times o}$; $\bm{V} \in \mathbb{R}^{N\times o}$ & EC embedding set; EHR embedding set\\
$\bm{v}^{(n)}_{i}$;$\bm{d}^{(n)}_{i}$;$\bm{M}^{(n)}_{i}$&the $i$-th visit; major diagnosis; treatment codes of patient $n$\\
$\bm{m}^{(n)}_{i,j}$&the $j$-th treatment code from $\bm{M}^{(n)}_{i}$\\
$\bm{U}_d;\bm{U}_m$& weight matrix for predicting $\bm{d}^{(n)}_{i}$ and $\bm{m}^{(n)}_{i,j}$\\
$\bm{W}_m;\bm{W}_o;\bm{W}_{p1};\bm{W}_{p2}$& weight matrix for MLP layer\\
$\bm{P}$ & set of demographics of given patient\\
$\bm{p}_k$ & the $k$-th demographics of given patient\\
$\bm{o}_i$ & visit-level embedding of $\bm{v}_i$\\
$\bm{u}$; $\bm{v}$ & EC embedding; EHR embedding\\
$\bm{\alpha}$; $\bm{\beta}$ & soft-algined phrases of $\bm{v}$ to $\bm{u}$; of $\bm{u}$ to $\bm{v}$ \\
$\bm{r_1}, \bm{r_2}$ & partly inference output from ($\bm{u}$, $\bm{\beta}$) and ($\bm{v}$, $\bm{\alpha}$)\\
$\bm{m}$ & aggregation output from $\bm{r_1}$ and $\bm{r_2}$\\
$f$;$g$;$h$;$r$ &  function for EHR hierarchical embeddings\\
$\sigma$ & non-linear activation function\\
$M$; $N$ & \# of EC statements; \# of EHR visits\\
${y, y'}$ &  real, predicted outcome \\
\bottomrule
\end{tabular}}
\end{table}

\subsection{The \mname Model}

\begin{figure*}[h!]
\centering
\includegraphics[scale=0.42]{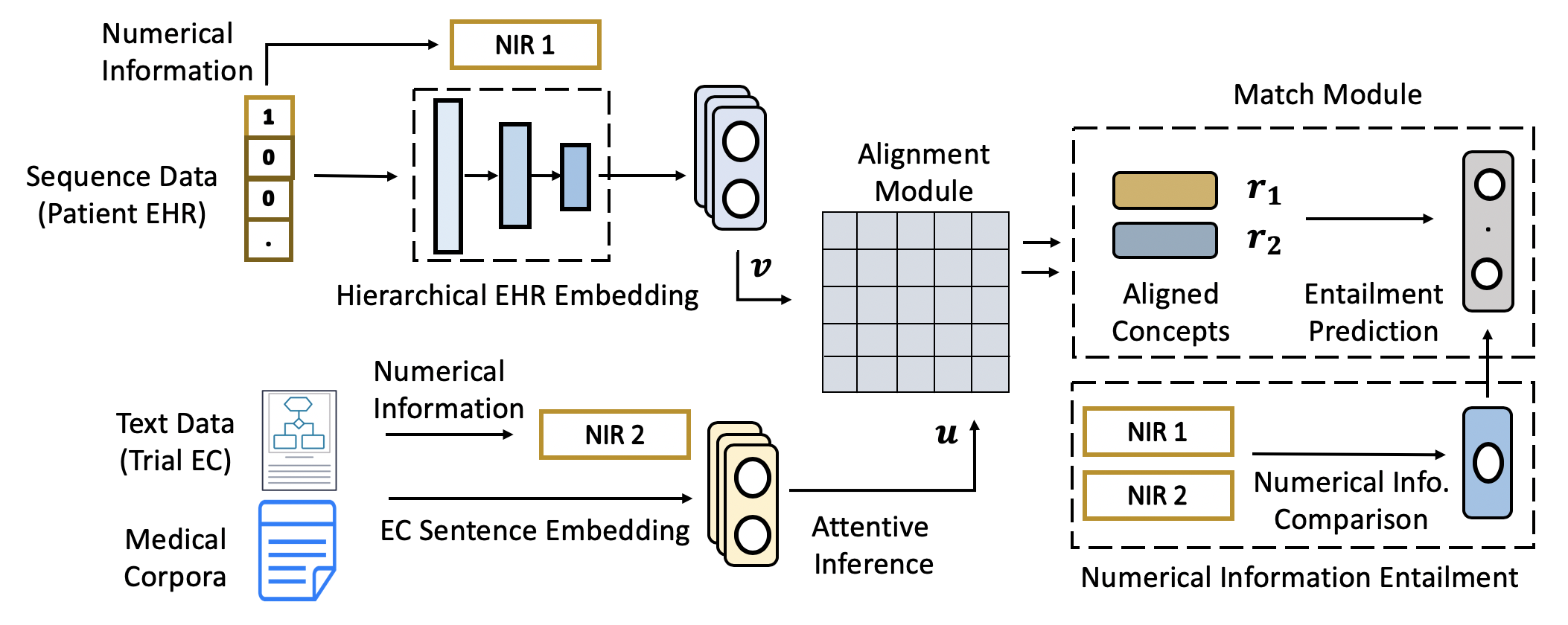}
\vskip -1em
\caption{The \mname framework. \mname applies a pre-trained BERT model to encode clinical trial information into sentence embedding $\bm{U}$. And the hierarchical embedding model encodes patient health data into embedding $\bm{V}$. Then the distributed representation $\bm{U}$ and $\bm{V}$ are soft-aligned to capture the interaction from each side. In addition, \mname is augmented by  a numerical information embedding and entailment module to explicitly match numerical information representations (NIR) and reason over numerical information in EC and EHR. At last, the match module predicts the matching scores from the interaction representations $\bm{r}_1$ and $\bm{r}_2$. The results will be further updated by the result of numerical information entailment.}
\vskip -1em
\label{fig1}
\end{figure*}

As illustrated in Fig.~\ref{fig1}, \mname includes the following components: a trial EC embedding module, a hierarchical patient representation module, the alignment and entailment prediction module that perform patient-trial matching via attentive comparison for soft-aligned fragments. Next we will first introduce these modules and then provide details of training and inference.\\

\noindent\textbf{Trial EC Sentence Embedding Using BERT}
Clinical trial ECs are in the form of unstructured text. To embed these data into meaningful vector representations, we use a pre-trained BERT model as the base model for word and sentence representations. Based on a multi-layer Transformer encoder~\cite{vaswani2017attention},
BERT is pre-trained using two unsupervised tasks.
\begin{enumerate}[leftmargin=*]
    \item Masked Language Model. Instead of predicting words based on previous words, BERT  predicts randomly masked words in a sequence to learn better bidirectional representations.
    \item Next Sentence Prediction. In order to gain better sentence embedding results which is used in our further matching prediction task, BERT has  a binary sentence prediction task to predict whether one sentence is the next sentence of the other.
\end{enumerate}
In particular, we applied the Clinical BERT~\cite{alsentzer2019publicly}, which was pre-trained on three medical corpora including PubMed abstracts, PubMed full-text articles, and MIMIC-III doctor notes. Then, we collect text data for clinical trials, including trial condition, summary, description and ECs to further fine-tune the Clinical BERT. Last, we use the fine-tuned BERT model to generate the EC sentence embedding for downstream patient-trial matching tasks.

The trial EC embedding process can be formally described below. We denote the fine-tuned BERT model as $d(\cdot)$ and the ECs of the clinical trial $n$ as $\bm{E}^{(n)} = (\bm{e}^{(n)}_1, \cdots \bm{e}^{(n)}_M)$, where each $\bm{e}^{(n)}_i$ indicates the $i$-th EC and is represented as a sequence of words: $\bm{e}^{(n)}_i=(\bm{a}^{(n)}_{i,1}, \cdots \bm{a}^{(n)}_{i,m})$. The embedding for each sentence is described as follow:
\begin{align}
   &\bm{u}^{(n)}_i=d(\bm{e}_i^{(n)})=d(\bm{a}^{(n)}_{i,1}, \cdots \bm{a}^{(n)}_{i,m})\\
   &\bm{U}^{(n)}=\{\bm{u}^{(n)}_1, \cdots \bm{u}^{(n)}_M\}
\end{align}
where $\bm{u}^{(n)}_i$ represents the sentence embedding of the $i$-th EC statement of clinical trial $n$ and $\bm{U}^{(n)}$ is the sequence of embedding for the entire clinical trial $i$. \\

\noindent\textbf{Hierarchical Patient Data Embedding}
Inspired by the multilevel patient embedding strategy in ~\cite{choi2018mime}, we also leverage the inherent multilevel structure of EHR data to learn patient embedding. The hierarchical structure of EHR data is structured as   patient-level, visit-level, then diagnosis codes and corresponding treatment codes for that visit. Demographics information is also provided in patient-level, including birth year, gender, country, geo location, ethnicity, and blood type. We encode these medical codes and demographics as using one-hot vectors. We first embedded into a dense representation, and then apply a hierarchical network to learn  patient embedding.

Formally, for each patient, we denote patient demographics as $\bm{p}_k \in \bm{P}$. Each visit $\bm{h}_i$ encompasses treatments $\bm{m}_{i,j} \in \bm{M}_{i}$ and diagnosis $\bm{d}_{i}$. The hierarchical embedding process is given below.
\begin{align}
   &g(\bm{d}_i,\bm{m}_{i,j})=\sigma(\bm{W}_m r(d_i))\odot r(\bm{m}_{i,j})\\
&f(\bm{d}_i,\bm{M}_i)=\bm{v}_i=\sigma(\bm{W}_o(r(\bm{d}_i)+\sum^{|\bm{M}_i|}_{j}g(\bm{d}_i,\bm{m}_{i,j})))\\
&h(\{\bm{p}\},\{\bm{v}\})=\sigma(\sum^{|\{\bm{p}\}|}_k\bm{W}_{\bm{p}_1} r(\bm{p}_k)+\sum^{|\{\bm{v}\}|}_i\bm{W}_{\bm{p}_2} \bm{v}_i)
\end{align}

We denote $r(\cdot)$ as a single MLP that encodes demographics data $\bm{p}_k$, diagnosis codes $\bm{d}_i$, and treatment codes $\bm{m}_{i,j}$ into a $z$-dimensional embedding. And a weighted matrix $\bm{W}_m \in \mathbb{R}^{z \times z}$ converts $r(\bm{d}_{i})$ into a different latent space to  effectively capture its interaction with $r(\bm{m}_{i,j})$. The function $f$ takes the distributed representation of diagnosis code $r(\bm{d}_i)$ and the sum of drug-treatment interactions to become the visit-level embedding. Then, the function $h$ will integrate demographics embedding and visit embedding into the patient-level embedding.

To fine-tune the patient embedding, we further develop an auxiliary prediction task to predict diagnosis code $\bm{d}_i$ and corresponding treatment codes $\bm{m}_{i,j}$  based on visit-level embedding $\bm{o}_i$.
\begin{align}
    &\bm{\overline{d}}_{i}=p(\bm{d}_i|\bm{o}_i)=softmax(\bm{U}_d\bm{o}_i)\\
    &\bm{\overline{m}}_{i,j}=p(\bm{m}_{i,j}|\bm{o}_i)=\sigma(\bm{U}_m\bm{o}_i)
\end{align}
Where $\bm{U}_d$ and $\bm{U}_m$ are weight matrix to compute the prediction of diagnosis code $\bm{\overline{d}}_{i}$ and treatment codes $\bm{\overline{m}}_{i,j}$ respectively.\\

\noindent\textbf{Alignment Module}
Attentive neural networks have demonstrated success in entailment prediction tasks ~\cite{parikh2016decomposable,rocktaschel2015reasoning}. In our case, we regard the trial EC embedding as premise and the encoded EHR as hypothesis. Our task is to predict whether the hypothesis is entailed in the premise, i.e. given a trial EC, whether a particular patient is a match for the trial.

First, we create a soft-alignment matrix using neural attention between a set of premises and hypotheses. Next, we use the alignment vectors to decompose our task into two sub-problems.
\begin{enumerate}[leftmargin=*]
    \item Given a certain premise, predict the entailment between the premise and all hypothesises.
    \item Given a certain hypothesis, predict the entailment between the hypothesis and all premises.
\end{enumerate}
Such a decomposition allows every EC to be soft-aligned to all related EHR, as well as every EHR to all related EC. Accordingly, the  match module can capture the entailment relations between all EC and EHR pairs.

Formally, we denote $\bm{U}=\{\bm{u}_1,\cdots, \bm{u}_M\}$ as a set of premises and $\bm{V}=\{\bm{v}_1,\cdots, \bm{v}_N\}$ as a set of hypothesises, where $\bm{u}_i,\bm{v}_j \in \mathbb{R}^{o}$. We feed elements from premise and hypothesis into a shared transformation layer parameterized by ($\bm{W}_c,\bm{b}_c$) and obtain unnormalized attention weights as soft-alignment matrix $\bm{W}_d$.
\begin{align}
    \bm{W}_d^{(i,j)} = \sigma(\bm{W}_c\bm{u}_i+\bm{b}_c)^T  \sigma(\bm{W}_c\bm{v}_j+\bm{b}_c).
\end{align}
Here $\sigma$ is an activation function (e.g., ReLU in our case) that operates element-wise over a vector. And $\bm{W}_d^{(i,j)}$ is the attention weight between $\bm{u}_i$ and $\bm{v}_j$. The soft-alignment matrix is then used to obtain the weighted summation for premise as $\bm{\beta}_i$ and hypothesis as $\bm{\alpha}_j$.
\begin{align}
    &\bm{\beta}_i = \sum_{j=1}\frac{exp(\bm{W}_d^{(i,j)})}{\sum_{k=1}exp(\bm{W}_d^{(k,j)})}\bm{v}_j, \\
  &\bm{\alpha}_j = \sum_{i=1}\frac{exp(\bm{W}_d^{(i,j)})}{\sum_{k=1}exp(\bm{W}_d^{(k,j)})}\bm{u}_i
\end{align}
Here $\bm{\beta}_i$ is the related hypothesis in $\bm{V}$ that is softly aligned to $\bm{u}_i$, while $\bm{\alpha}$ is the related premise in EC $\bm{U}$ that is softly aligned to $\bm{v}_j$. And $\{\bm{\beta}_i\}$ is formal representation for the input of sub-problems (1). $\{\bm{\alpha}_j\}$ is formal representation for the input of sub-problems (2).

The proposed decomposable soft-alignment has two major benefits. First, the aligned vectors focus on only related local sub-structure, which is proved to be effective for entailment inference task~\cite{parikh2016decomposable}. Second, the decomposition avoids the quadratic complexity ($M \times N$ times) and only requires linear complexity ($M + N$ times).\\

\noindent\textbf{Match  via Entailment Prediction}
Given the output from the alignment module, we jointly consider interactions between single EHR and EC as well as for the original EC embeddings and EHR embeddings $(\bm{v}_j,\bm{\alpha}_j)$  and $(\bm{u}_i,\bm{\beta}_i)$ by feeding their concatenation into a shared transformation layer to compute the entailment relationship between given patient and trial.
\begin{align}
    & \bm{r}_{1,i} = \sigma(\bm{W}_a[(\bm{u}_i,\bm{\beta}_i)] + \bm{b}_a)\\
    & \bm{r}_{2,j} = \sigma(\bm{W}_a[(\bm{v}_j,\bm{\alpha}_j)] + \bm{b}_a)
\end{align}
Where ($\bm{W}_a$,$\bm{b}_a$) are the parameters for the transformation layer. The output $\bm{r}_{1,i}$ and $\bm{r}_{2,j}$  represent the entailment prediction results for sub-problems (1) and sub-problems (2), respectively. They are used to aggregate into the final entailment inference results.
\begin{align}
    & \bm{r}_{1}=\sum_{i=1}{\bm{r}_{1,i}}, \quad \bm{r}_{2}=\sum_{j=1}\bm{r}_{2,j}
\end{align}
Where $\bm{r}_{1}$ and $\bm{r}_{2}$ are the summation of the prediction results of the two sub-problems. Following the aggregation method from \cite{mou2015natural}, we adapt three aggregation methods:
(a) vector concatenation;(b) element-wise product; (c) element-wise difference, to measure their similarity and closeness in latent space. The aggregated representations are then feed into a final output layer.
\begin{align}
    & \bm{m}= [\bm{r}_{1} , \bm{r}_{2} , \bm{r}_{1}\cdot\bm{r}_{2} , \bm{r}_{1}-\bm{r}_{2}]\\
    & y' = p(y|\bm{m}) = \mathrm{softmax}(\bm{W}_f\bm{m}+\bm{b}_f)
\end{align}
where ($\bm{W}_f$,$\bm{b}_f$) are the parameters for output layer. $y'$ represents the predicted label, including "entailment", "contradiction", and "neutral". The entire embedding and entailment prediction process is summarized in Algorithm 1.
\\

\noindent\textbf{Numerical Information Entailment}
Numerical information in ECs such as age or dosage is important for patient-trial matching. However, it is difficult to match them since  numbers are a common source of contradictions in natural language processing tasks~\cite{dagan2013recognizing}, and are often insensitive in distributed representations~\cite{roy2015reasoning}.
To explicitly model these numerical information, we proposed the following {\it numerical information representation}(NIR).

The design of NIR is inspired by~\cite{roy2015reasoning,ho2019qsearch}. We first encode numerical information as a triplet ($q,u,c$), where the three variables  correspond to number, unit, and concept, respectively.
\begin{itemize}[leftmargin=*]
\item Number $q$: The lower or upper bound for a value range. e.g. more than 500 mg, at least one month, during last three months. We do not store the surface forms but convert them into a set of ranges. For example, "more than 20 mg" is stored as (20,+$\infty$).
\item Unit $u$: is about the scale a value is measured. e.g., mg, weeks. The word "weeks" in the phrase "within 12 weeks" is a unit.
\item Concept $c$: an aligned concept that the quantity is associated with. e.g. certain drug or procedure. This is stored for augmentation of quantity and input representations.
\end{itemize}

Next, we develop the following method.
Given a sequence of tokens $\{\bm{x}_i\}$, we extract its word-level features and character-level features to recognize numerical information. In word-level, we can recognize if $\bm{x}_i$ appears in a set of known scientific units(e.g., mcg, g/L), written numbers. In character-level, $\bm{x}_i$ can contain a digit, have all digits, or include a suffix (st,nd,rd,th). Then the extraction results are formulated as the numerical information representation with standardization including numbers converting into floating points and fixed date type, and units converting into standard base unit.

Finally, we design a numerical information comparison method to infer the entailment between NIR learned from ECs and numerical information in EHR. Numerical information in EHR are often represented as a fixed value with unit. Therefore, we firstly check whether two units are comparable. If comparable, we would then compare the fixed value with the value range in NIR. The entailment results would includes: "entailment", "contradiction" and "not comparable". Only the "entailment" ones should be regarded as supported by the numerical information entailment module.

\begin{algorithm}[!h]
    \label{alg1}
	\caption{\mname for Patient-Trial Matching }
	\hspace*{0.02in} {\bf Input:} A set of labelled triplet $\{\bm{E}^{(i)}, \bm{H}^{(i)}, y_i\}_{i=1}^{i=N}$. Let $N$ equals the size of training set.
	\begin{algorithmic}[1]
		\FOR{iter $= 1\ldots N$}
		\STATE $\bm{U}^{(i)}$ $\leftarrow$ EC Sentence Embedding($\bm{E}^{(i)}$)
		\STATE $\bm{V}^{(i)}$ $\leftarrow$ Hierarchical Patient Embedding($\bm{H}^{(i)}$)
		\STATE $\{\bm{\beta}_j\}$,$\{\bm{\alpha}_k\}$ $\leftarrow$ Concept Alignment($\bm{U}^{(i)}, \bm{V}^{(i)}$)
		\IF{Entailment Match($\{\bm{\beta}_j\}$,$\{\bm{\alpha}_k\}$) = {\it Entailment}}
		\IF{{Quantity Match($\bm{E}^{(i)}, \bm{H}^{(i)}$) = {\it Entailment}}}
		\RETURN \textit{Entailment}
		\ELSE
		\RETURN \textit{Contradiction}
		\ENDIF
		\ELSIF {Entailment Match ($\{\bm{\beta}_j\}$,$\{\bm{\alpha}_k\}$) = {\it Contradiction}}
		\RETURN \textit{Contradiction}
		\ELSIF {Entailment Match ($\{\bm{\beta}_j\}$,$\{\bm{\alpha}_k\}$) = {\it Neutral}}
		\RETURN \textit{Neutral}
		\ENDIF
		\ENDFOR
	\end{algorithmic}
	\hspace*{0.02in} {\bf Output:}
	Trained deep embedding and entailment prediction model.
\end{algorithm}

\section{Experiment}

We evaluated \mname  model on three datasets, including  two sets of proprietary matched trial and EHR data, and a publicly available synthetic data.  The code can be found in~\footnote{https://github.com/deepenroll/DeepEnroll} We designed experiments to answer the following questions.

\noindent \textbf{Q1}: Does \mname have better performance in matching patients and trials as measured by predicting clinical trial enrollment?

\noindent \textbf{Q2}: Does the numerical information entailment module help improve the performance of \mname ?

\noindent \textbf{Q3}: Can \mname provide interpretable matching results?

\subsection{Experimental Setting}

\noindent\textbf{Data}
The data used in experiments are listed below.
\begin{itemize}[leftmargin=*]
    \item \textbf{Clinical Trial EC}
    We randomly selected 794 clinical trials with varying disease domains from ClinicalTrials.gov. Firstly we downloaded the XML files for the clinical trials. Each file follows a fixed structure defined by the clinical trial XML schema. We select the \textit{Inclusive Criteria} and \textit{Exclusive Criteria} to form raw ECs. Then we extract EC statements from the free-text raw ECs through patter-based paragraph segmentation and sentence segmentation. In total, we obtain 12,445 sentence-level EC statements for 794 clinical trials.
    \item \textbf{Large Scale Patient-Trial Matching Data (IQVIA dataset)}
    We use this trial-patient matched data for model training and validation, noted as IQVIA dataset. It contains 561 registered clinical trials and 57,696 matched patients. The patient information is encoded in a large-scale longitudinal prescription and medical claims data.
    For each trial, a number of free-text EC statements describe trial requirements from different aspects. Patients are profiled as a set of EHR, including records of diagnosis, and treatments. We define the inclusive EC and exclusive EC and their corresponding matched patients' EHR as one "Entailment" or "Contradiction". As to unmatched EHR and EC statements, we labelled them as "Neutral". In all we have 852,1031 labelled examples.
    \item \textbf{Rare Disease Data}. The rare disease data is a subset of IQVIA dataset, which contains conditions from the rare diseases list provided by the national organization for rare disorders(NORD). This results in 77 trials and 562 matched patients from IQVIA dataset.
    \item \textbf{Synthetic Data} For reproducibility, we also develop a synthetic data based on~\cite{yuan2019criteria2query} using Synthea~\footnote{https://github.com/synthetichealth/synthea}, a synthetic patient population simulator to automatically generate patient EHR. We select 243 registered trials from a wide range of disease domains, and leverage Criteria2Query to generate cohort definitions for filtering matched patient EHRs. In total, we have 68,845 simulation EHRs, which is labelled in similar way as IQVIA dataset. Details for synthetic data generation can be found in Section~\ref{sec:appendixdetail} of the appendix.
\end{itemize}

\begin{table} [h!]
\caption{Data Statistics}
\centering
\begin{tabular}{lccc}
 \toprule
\textbf{Statistic}& IQVIA & Synthetic & Rare Disease  \\
\midrule
\# of trials & 561 & 243 & 77\\
 \# of EC statements& 9,510 & 2,935 & 1,394 \\
Avg. EC sentence length & 24.2 & 21.3 & 23.4 \\
\# of patients & 57,696 & 13,634 & 562 \\
\# of EHR & 852,103 & 68,845 & 9,580 \\
\# of unique medications& 352 & 156 & 294 \\
\# of unique diagnosis& 939 & 288 & 309\\
\bottomrule
\end{tabular}
\end{table}



\noindent\textbf{Baseline}
We compared \mname with the following baselines.
\begin{itemize}[leftmargin=*]
\item \noindent\textbf{Logistic Regression (LR)}~\cite{hosmer2013applied}
We concatenate sentence embedding of EC and one-hot EHR vectors and apply LR to make multiclass classification.

\item \noindent\textbf{Multi-Layer Perception (MLP)} is leveraged as entailment inference module~\cite{bowman2015large,conneau2017supervised}. The model is simply a stack of three 200d tanh layers, with the bottom layer taking the concatenated premise and hypothesis word representations as input and the top layer feeding a softmax classifier.
\item \noindent\textbf{Long Short Term Memory (LSTM)}~\cite{hochreiter1997long}. We apply two LSTM to model both premise sequence and hypothesis sequence and then use the corresponding final output for entailment prediction.
\item \noindent\textbf{match-LSTM (mLSTM)}~\cite{wang2015learning} perform word-by-word matching of the hypothesis with the premise.
\item \noindent\textbf{Stack Augmented Parser Interpreter Network (SPINN)}~\cite{bowman2016fast} combines parsing
and interpretation within a tree-sequence hybrid model by integrating tree structured sentence interpretation.
\item \noindent\textbf{Word-by-word Attention(WA)}~\cite{rocktaschel2015reasoning} is an entailment inference model that utilize the output latent state of premise LSTM as the input of hypothesis LSTM. Then a word-by-word attention is developed for both LSTM for entailment prediction.


\item \noindent\textbf{Criteria2Query}~\cite{yuan2019criteria2query} is the current state-of-the-art work for patient-trial matching task. It implements a systematic information extraction  pipeline to parse free-text eligibility criteria into a structured and computable representation using the OMOP CDM, and also leverages a series of natural language inference techniques such as named entity recognition to autonomously generate matching patient definition to identify patient cohorts Note that we would not compare its performance on Synthea dataset, which is generated through Criteria2Query.
\end{itemize}

\noindent\textbf{Metrics}
To measure the accuracy of matching, we use the average F1 and Precision Recall AUC (PR-AUC) as our  metrics.
Precision, Recall, and F1 score are computed on the number of true positives (TP), false positives (FP), and false negatives (FN). PR-AUC measures the area under the precision-recall curve. F1 score is the harmonic mean of precision and recall. Since the entailment prediction task is casted as a multi-label classification task, we consider micro-F1 as the metric and calculate average F1 based on that. \\

\noindent\textbf{Evaluation Strategy} All methods are implemented in PyTorch~\cite{paszke2017automatic} and trained on an Ubuntu 16.04 with 64GB memory and three GTX 1080 Ti GPU. For all datasets, we randomly select 60\% of the patients as training set, 20\% as validation set and the remaining 20\% as test set. We train the model using training data, and fix model parameters based on the best model performance on validation set. We then test the model on test set. We perform three random runs and report both mean and standard deviation for testing performance.
\\

\noindent\textbf{Implementation Details}.
We use stochastic gradient descent (SGD) with a learning rate of 0.1 and a weight decay of 0.99. At each epoch, we divide the learning rate by 5 if the dev accuracy decreases. We use mini-batches of size 64 and training is stopped when the learning rate goes under the threshold of 10e-5. For the classifier, we use a multi-layer perception with 1 hidden-layer of 512 hidden units. We use dropout~\cite{srivastava2014dropout} with a rate of 0.5, which is applied to all feedforward connections. For the pre-trained Clinical BERT, We use a batch size of 32 and fine-tune for 3 epochs over the data for two unsupervised task. For each task, we selected the fine-tuning learning rate of 2e-5.

For all baselines except for Criteria2Query, we use 300 dimensional GloVe embedding~\cite{Pennington14glove:global} as word embedding and max-pooling as sentence embedding method. Out-of-vocabulary (OOV) words are hashed to one of 100 random embedding each initialized to mean 0 and standard deviation 1. All other hidden layer weights were initialized from random Gaussian distribution with mean 0 and standard deviation 0.01. Each hyperparameter setting was run on a same machine as the \mname, using Adagrad~\cite{Duchi:2011:ASM:1953048.2021068} for optimization with initial accumulator value of 0.1. For EHR data, we convert them into one-hot vectors as the hypothesis representation. As for Criteria2Query, we directly use the existing model to produce the matching results for given trials and patients.

\subsection*{Q1: \mname achieved the best predictive performance in clinical trial enrollment prediction}

We compared \mname against the state-of-the-art baselines on IQVIA dataset, the synthetic data, and the rare disease data. We reported the results in Table.~\ref{table:resultiqvia}, Table~\ref{table:syntheticresult} and Table~\ref{table:resultrare}, respectively. The best results are presented in bold figures.

\begin{table} [h!]
\caption{Performance Comparison on IQVIA dataset.}~\label{table:resultiqvia}
\centering
\resizebox{\columnwidth}{!}{
\begin{tabular}{lcc}
 \toprule
Model & Average F1& PR-AUC \\  \midrule
LR & 0.5675$\pm$ 0.0003& 0.6580$\pm$ 0.0005\\
MLP& 0.5864$\pm$ 0.0001& 0.6738$\pm$ 0.0000\\
LSTM& 0.6225$\pm$0.0001& 0.6871$\pm$0.0001\\
SPINN& 0.6331$\pm$0.0042& 0.7014$\pm$0.0063\\
Word-by-word Attention& 0.6457$\pm$0.0012& 0.7163$\pm$0.0021\\
match-LSTM& 0.6641$\pm$0.0007& 0.7423$\pm$0.0011\\
Criteria2Query& 0.6819$\pm$0.0001& 0.7103$\pm$0.0002\\
\bf \mname&  \bf 0.7543$\pm$0.0018& \bf 0.7864$\pm$0.0083\\
\bottomrule
\end{tabular} }
\end{table}

\begin{table} [h!]
\caption{Performance Comparison on synthetic data. C2Q is not compared here since it was used in  data synthesis.}~\label{table:syntheticresult}
\centering
\resizebox{\columnwidth}{!}{
\begin{tabular}{lcc}
 \toprule
Model & Average F1& PR-AUC \\  \midrule
LR & 0.6123$\pm$0.0002& 0.6849$\pm$0.0003\\
MLP& 0.6231$\pm$0.0001& 0.6914$\pm$0.0000\\
LSTM& 0.6424$\pm$0.0002& 0.7092$\pm$0.0003\\
SPINN& 0.6233$\pm$0.0074& 0.6712$\pm$0.0078\\
Word-by-word Attention& 0.6245$\pm$0.0019& 0.6821$\pm$0.0034\\
match-LSTM& 0.6542$\pm$0.0013& 0.6956$\pm$0.0025\\
\bf \mname&\bf 0.7298$\pm$0.0013& \bf 0.7431$\pm$0.0029\\
\bottomrule
\end{tabular} }
\end{table}

\begin{table} [h!]
\label{rareDisease}
\caption{Performance comparison on rare disease data.
}~\label{table:resultrare}
\centering
\begin{tabular}{lcc}
\toprule
 Model & Average F1& PR-AUC \\
\midrule
LR & 0.5518$\pm$ 0.0004& 0.6375$\pm$ 0.0012\\
MLP& 0.5454$\pm$ 0.0001& 0.6844$\pm$ 0.0000\\
LSTM& 0.6163$\pm$0.0002& 0.6782$\pm$0.0002\\
SPINN& 0.6194$\pm$0.0061& 0.6892$\pm$0.0078\\
Word-by-word Attention& 0.6357$\pm$0.0014& 0.7065$\pm$0.0023\\
match-LSTM& 0.6442$\pm$0.0009& 0.7133$\pm$0.0021\\
Criteria2Query& 0.6155$\pm$0.0001& 0.6481$\pm$0.0001\\
\bf \mname& \bf 0.7241$\pm$0.0011&\bf 0.7453$\pm$0.0029\\
\bottomrule

\end{tabular}
\end{table}

From the results, we observe that \mname consistently achieves the best performance against best baselines. On IQVIA dataset, it outperformed the best baseline Criteria2Query by 10.6\% in Average F1 and match-LSTM by 5.9\% in PR-AUC. On the synthetic data, it outperformed the best baseline match-LSTM by 11.5\% in Average F1 and 6.8\% in PR-AUC. On rare disease dataset, it outperformed the best baseline match-LSTM by 12.4\% in Average F1 and 4.5\% in PR-AUC.

Among the baselines, LR and MLP have the worst performance due to they fail to capture the temporal patterns in EC text and patient data, thus cause the inferior matching performance. Compared with them, the LSTM and the word-by-word attention model manage to capture long-range temporal dependencies among both EC and patient data. The SPINN model uses a tree-sequence hybrid structure for better sequence modeling. Thus they demomnstrated improved performance over LR and MLP.

The two best baselines are match-LSTM and Criteria2Query. The match-LSTM model is a LSTM-based natural language inference model.  It performs word-by-word matching of the hypothesis with the premise, such that it not only can place more emphasis on important word-level matching results, but also can remember important mismatches that are critical for predicting the contradiction or the neutral relationship label. With this design, match-LSTM has greatly improved performance compared with the other LSTM-based models. On the other hand, Criteria2Query~\cite{yuan2019criteria2query} implements a systematic information extraction  pipeline to parse free-text eligibility criteria into a structured and computable representation using the OMOP CDM, and also leverages a series of natural language inference techniques such as named entity recognition to autonomously generate matching patient definition to identify patient cohorts. It achieved very good performances across all clinical trials due to rule extraction and natural language inference techniques. However, on rare disease trial data, the performance of Criteria2Query is not encouraging, which could be due to the lack of full knowledge of rare diseases in the knowledge system it builds on. One example was that Criteria2Query showed confusing results in recognize numerical restrictions: Heart failure with ejection fraction $\le$
40 (NCT04078425). Criteria2Query would recognize only Heart failure while DeepEnroll could recognize the specific numeric criteria for matching.

To compare with retrieval-style baseline, we used solr\cite{smiley2011apache} to query 20 clinical trials with textual descriptions of EHR formatted in bag of words, including demographics and name of diagnosis, products, and procedures. The F-1 score for top 5 results is 0.4731 and for top 10 is 0.5378 (vs. DeepEnroll 0.7543). As to the baselines mentioned in Related Work, the baseline Criteria2Query already uses the same retrieval component as EliXR and EliIE.

In this experiment, we also evaluated model performance on patient-trial matching for rare diseases. Recruiting suitable patients to trials for rare diseases is challenging. This is because often there are few patients with rare diseases. It may not be feasible to significantly narrow entry criteria based on disease stage or other characteristics~\cite{Augustine2013}. From the results in Table~\ref{table:resultrare}, we can observe that although the performance of all models decreased, \mname still output performed all baselines with a 7.99\% improvement and demonstrated a minimal performance reduction. Among  baselines, Criteria2Query has the biggest performance drop since it heavily relies on disease guidelines, which are not sufficient for rare diseases.

\subsection*{Q2: The numerical information entailment module augments the performance of \mname}

We conduct an ablation study to understand the contribution of numerical information representation and entailment module (NIR) in \mname. We remove the NIR module in patient-trial matching and perform trial enrollment prediction using the reduced model. We compared the prediction results against the full model across all datasets. The parameters in the reduced model are determined with cross-validation, and the best performances are reported in Table \ref{tab:ablation}.

\begin{table}[h!]
\caption{Abalation study of \mname demonstrated the advantage of numerical information entailment augmentation.}
\label{tab:ablation}
\centering
\resizebox{\columnwidth}{!}{
\begin{tabular}{lccc}
\toprule
Dataset & Metric & w/o NIR  &  \mname \\ \hline
\multirow{2}{*}{IQVIA dataset} & PR AUC & 0.7611$\pm$0.0094  & \bf 0.7864$\pm$0.0083 \\
 & F1 Score & 0.7392$\pm$0.0021 & \bf 0.7543$\pm$0.0018 \\
 \hline
 \multirow{2}{*}{Rare Disease} & PR AUC & 0.7258$\pm$0.0032 & \bf 0.7453$\pm$0.0029 \\
 & F1 Score & 0.7012$\pm$0.0011  & \bf 0.7241$\pm$0.0011 \\
 \hline
\multirow{2}{*}{Synthetic Data} & PR AUC & 0.7399$\pm$0.0028  & \bf 0.7431$\pm$0.0029 \\
 & F1 Score & 0.7211$\pm$0.0013 & \bf 0.7298$\pm$0.0013 \\
 \bottomrule
\end{tabular}}
\end{table}

From Table.~\ref{tab:ablation}, it is easy to see when we solely leverage the EC sentence embedding and hierarchical patient embedding for entailment match, the performances are largely reduced. For example, on IQVIA dataset and rare disease data, the performance could be decreased by up to 3.32\%  in PR-AUC and 3.26\% in F1 score.  In conclusion, it suggests the necessity of augmenting textual representation with explicit embedding and entailment for numerical information in trial-patient matching task.

\subsection*{Q3:  \mname can provide interpretable matching results}
To better facilitate clinical decision making, we also design the following strategy to understand the key factors for matching patients and trials.
 For each trial considered in case study, we randomly select $12$ patients who are enrolled in the trial.  We collect the soft-alignment attention weights, and then visualize them using heatmaps.

 Here, we select the following trials: trials on heart failure (HF, NCT03882710), Alzheimer's Disease (AD, NCT03690193) and  idiopathic pulmonary fibrosis (IPF, NCT02085018).  Each row represent one EC statement and one column shows one patient embedding. We chose 5-6 representative EC statements with relatively high attention weights. We leveraged the weighted average of attention weights between EHR and EC into patient level ($\bm{\beta}$), and visualize the impact of each EC statements on the selected patients. We utilize ECharts as our visualization tool~\footnote{https://www.echartsjs.com}. Light color indicates lower relevance while dark color means strong relevance.

\begin{figure}[h!]
\includegraphics[scale=0.4]{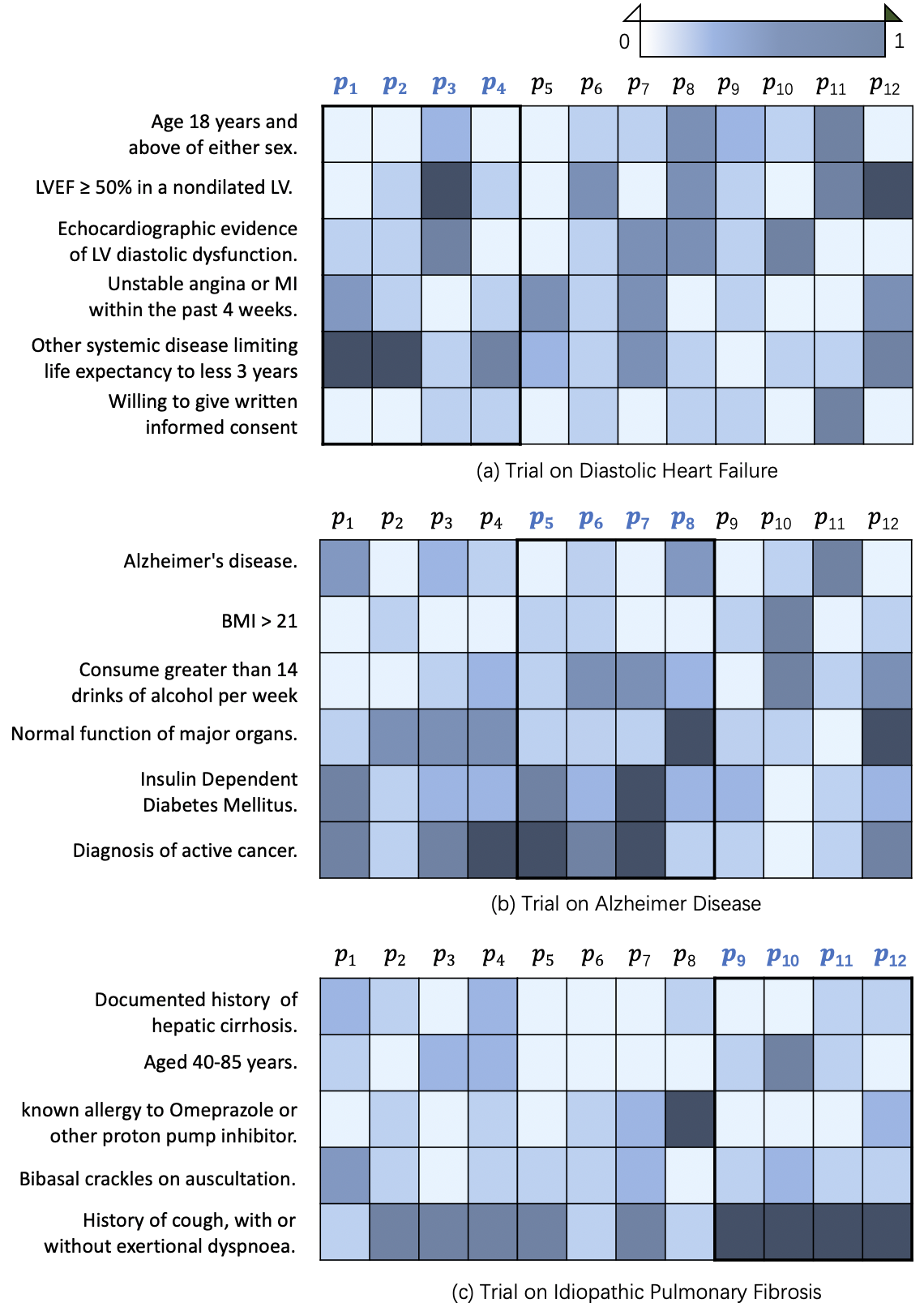}
\vskip -1em
\caption{Heatmap showing the attention weights for EC statements and patient pairs.  Rows represent EC statements (e.g., inclusion and exclusion) and columns are patient embedding. Dark color indicates strong relevance. Matched patients of each trial are highlighted.}\label{fig:heatmap3}
\vskip -1em
\end{figure}

Heart failure is a serious condition such that the heart of HF patient  cannot keep up with its workload. The body of the patient may not get the oxygen it needs. HF has no cure. Drugs for HF are designed to help patients better manage their conditions. From Figure~\ref{fig:heatmap3} (a), the most relevant EC statement requires the patients to have no "other systemic disease limiting life expectancy to less than 3 years" mainly due to the the clinical trial will take more than 3 years. Another important statement is about "LVEF $\ge 50\%$ in a nondilated LV", which is a key sign of HF.

Alzheimer's disease (AD) is an irreversible, progressive brain disorder that slowly destroys memory and thinking skills, and, eventually, the ability to carry out the simplest tasks. Currently, there is no cure for AD  or a way to stop its progression. Drugs and treatments are developed to either cure AD or help treat some symptoms of AD.  From Figure~\ref{fig:heatmap3} (b), the most relevant EC statement states that in order for the patients to be considered for the trial, their major organ should have normal functions, and also  patients with diabetes and cancer should be excluded.  For AD trial,  BMI measure and behavior description are less important.

The last example is for a trial on treating idiopathic pulmonary fibrosis (IPF). IPF is a pulmonary disease that is characterized by the formation of scar tissue within the lungs in the absence of any known provocation~\cite{Meltzer2008}. IPF is a rare disease which affects approximately 5 million people worldwide, with prevalence rate at $0.04\%$. Currently there is no cure for IPF.  Treatments are being developed and tested to slow the rate of scarring (pirfenidone and nintedanib) and treat particular IPF symptoms such as breathlessness and coughing. From Figure~\ref{fig:heatmap3} (c), for this trial the most relevant EC statement states the patients should have "history of cough", and have no "known allergy to Opeprazole or other proton pump inhibitor".\\

Through visualization, we conclude that the attention weights in \mname help better understand the key inclusion and exclusion criteria for successful recruitment of patients.

\subsection{Case Study: trials that are difficult to match patients with machine learning models}

\begin{figure}[h!]
\includegraphics[scale=0.38]{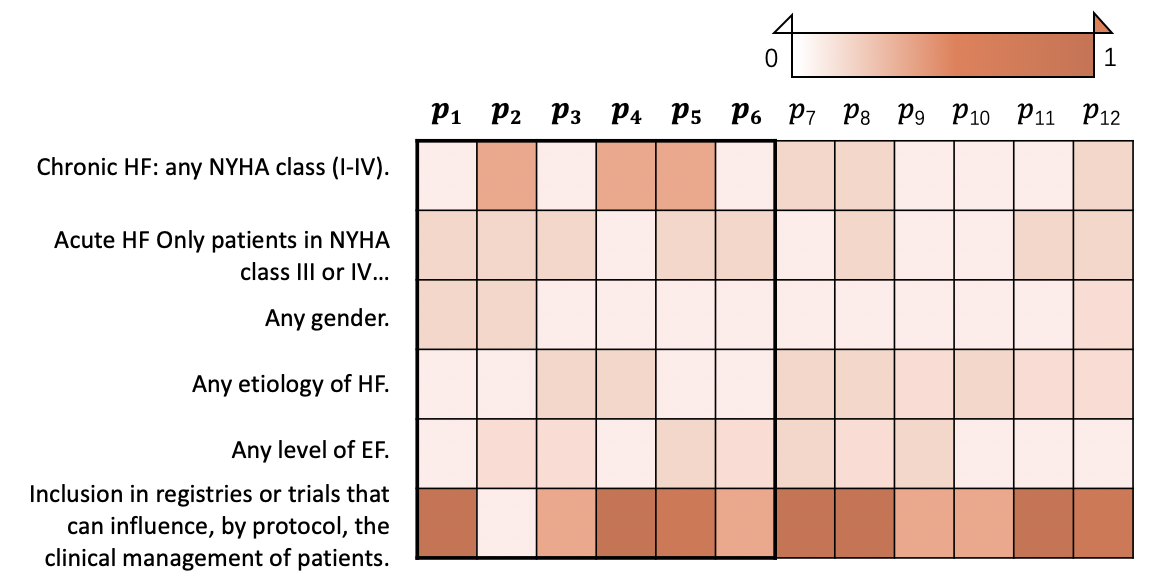}
\vskip -1em
\caption{Attention weights for BLITZ Heart Failure trial. Matched patients are highlighted.}\label{fig:inference_study}
\vskip -1em
\end{figure}
Among our experiments, for some clinical trials, it is not easy to use machine learning models to match patients. To analyze the possible reason, we select a trial that has shown lower than 30\% correct patient-trial matching using either \mname or Criteria2Query. The trial is named BLITZ Heart Failure (NCT03661060), which is an observational study designed to evaluate whether a structured educational program can improved the patient adherence from both acute and chronic HF patients to guidelines recommendations.

We take a similar strategy to visualize its attention matrix in Figure~\ref{fig:inference_study}. There are two possible reasons for the low matching performance. First, there are many over-simplified abbreviation in trial EC statements, such as the ones shown in the forth and fifth rows. The over-simplified description could be ambiguous and hard to match with concepts in patient EHR. In addition, the EC statement in the sixth row has relatively high attention weights, but is actually unrelated to information from patient EHR.

\section{Conclusion}
In this paper,  we proposed \mname,  a cross-modal inference learning model to jointly encode enrollment criteria  and patients records  into a shared latent space for patient-trial matching. \mname applies a pre-trained BERT model to encode clinical trial information, and a hierarchical model to embed patient EHR.  \mname is also augmented by a numerical information embedding and entailment module  to explicitly match numerical information in EC and EHR.
We eva both real world  clinical trial dataset and a synthetic data. Experiments on real-world datasets demonstrated the strong performance of \mname.
The \mname method can also be extended to other application domains for matching problems based on heterogeneous data, such as personalized targeted advertising that is based on matching Ads description (in text) and customers' browsing history (in sequential data).

\section{Acknowledgments}
This work is part supported by National Science Foundation award IIS-1418511, CCF-1533768 and IIS-1838042, the National Institute of Health award NIH R01 1R01NS107291-01 and R56HL138415.

\section{Appendix: Details on generating the synthetic dataset}~\label{sec:appendixdetail}
For model reproducibility, we generated a synthetic data following the procedure below.
\begin{enumerate}[leftmargin=*]
\item  \textit{Data Gathering}.
For clinical trial data, we downloaded  1,000 clinical trial descriptions from a  https://clinicaltrials.gov. Each trial is stored in an XML file that follows a structure of fields defined by an XML schema. We select the EC items in free-text format from XML files as our original dataset. Then, we split EC into statements by considering different types of bullet and lists symbols as the natural sentence splitting signs.
\item \textit{EHR Generation}.
According to \cite{weng2010formal}, EC items could be divided into the following categories:demographic information; condition occurrence, procedure occurrence; measurement; drug exposure; observation; and willing information. Except for willing and demographic information, other categories can be mapped with corresponding tables in EHR system.
For each EC record, we use the Criteria2Query~\cite{yuan2019criteria2query} system to generate a corresponding EHR query in CDM v5.0 format, which includes the category label for EC and the EHR query information.
Then, we run these queries in ATLAS (github.com/OHDSI/Atlas), an online EHR query engine to generate eligible patient-level EHR information. These information are stored in relational database. We developed a algorithm to translate the database information into natural text.

\item \textit{Labelling}.
Last, the matching EHR  and EC are labelled as Entailment for inclusion criteria and Contradiction for exclusion criteria. Additionally, a number of unrelated EHR and EC pairs are also generated and labelled as Neutral. The labeling process forms tuples in the form of $(\bm{E},\bm{H},y)$ as model input.
\end{enumerate}

\bibliographystyle{ACM-Reference-Format}
\bibliography{_manuscript.bib}


\begin{thebibliography}{36}


\ifx \showCODEN    \undefined \def \showCODEN     #1{\unskip}     \fi
\ifx \showDOI      \undefined \def \showDOI       #1{#1}\fi
\ifx \showISBNx    \undefined \def \showISBNx     #1{\unskip}     \fi
\ifx \showISBNxiii \undefined \def \showISBNxiii  #1{\unskip}     \fi
\ifx \showISSN     \undefined \def \showISSN      #1{\unskip}     \fi
\ifx \showLCCN     \undefined \def \showLCCN      #1{\unskip}     \fi
\ifx \shownote     \undefined \def \shownote      #1{#1}          \fi
\ifx \showarticletitle \undefined \def \showarticletitle #1{#1}   \fi
\ifx \showURL      \undefined \def \showURL       {\relax}        \fi
\providecommand\bibfield[2]{#2}
\providecommand\bibinfo[2]{#2}
\providecommand\natexlab[1]{#1}
\providecommand\showeprint[2][]{arXiv:#2}

\bibitem[\protect\citeauthoryear{Alicante, Corazza, Isgr{\`o}, and
  Silvestri}{Alicante et~al\mbox{.}}{2016}]%
        {alicante2016unsupervised}
\bibfield{author}{\bibinfo{person}{Anita Alicante}, \bibinfo{person}{Anna
  Corazza}, \bibinfo{person}{Francesco Isgr{\`o}}, {and}
  \bibinfo{person}{Stefano Silvestri}.} \bibinfo{year}{2016}\natexlab{}.
\newblock \showarticletitle{Unsupervised entity and relation extraction from
  clinical records in Italian}.
\newblock \bibinfo{journal}{\emph{Computers in biology and medicine}}
  \bibinfo{volume}{72} (\bibinfo{year}{2016}), \bibinfo{pages}{263--275}.
\newblock


\bibitem[\protect\citeauthoryear{Alsentzer, Murphy, Boag, Weng, Jin, Naumann,
  and McDermott}{Alsentzer et~al\mbox{.}}{2019}]%
        {alsentzer2019publicly}
\bibfield{author}{\bibinfo{person}{Emily Alsentzer}, \bibinfo{person}{John~R
  Murphy}, \bibinfo{person}{Willie Boag}, \bibinfo{person}{Wei-Hung Weng},
  \bibinfo{person}{Di Jin}, \bibinfo{person}{Tristan Naumann}, {and}
  \bibinfo{person}{Matthew McDermott}.} \bibinfo{year}{2019}\natexlab{}.
\newblock \showarticletitle{Publicly available clinical BERT embeddings}.
\newblock \bibinfo{journal}{\emph{arXiv preprint arXiv:1904.03323}}
  (\bibinfo{year}{2019}).
\newblock


\bibitem[\protect\citeauthoryear{Augustine, Adams, and Mink}{Augustine
  et~al\mbox{.}}{2013}]%
        {Augustine2013}
\bibfield{author}{\bibinfo{person}{EF. Augustine}, \bibinfo{person}{HR. Adams},
  {and} \bibinfo{person}{JW. Mink}.} \bibinfo{year}{2013}\natexlab{}.
\newblock \showarticletitle{Clinical trials in rare disease: challenges and
  opportunities}.
\newblock \bibinfo{journal}{\emph{J Child Neurol.}} (\bibinfo{year}{2013}).
\newblock


\bibitem[\protect\citeauthoryear{Bowman, Angeli, Potts, and Manning}{Bowman
  et~al\mbox{.}}{2015}]%
        {bowman2015large}
\bibfield{author}{\bibinfo{person}{Samuel~R Bowman}, \bibinfo{person}{Gabor
  Angeli}, \bibinfo{person}{Christopher Potts}, {and}
  \bibinfo{person}{Christopher~D Manning}.} \bibinfo{year}{2015}\natexlab{}.
\newblock \showarticletitle{A large annotated corpus for learning natural
  language inference}.
\newblock \bibinfo{journal}{\emph{arXiv preprint arXiv:1508.05326}}
  (\bibinfo{year}{2015}).
\newblock


\bibitem[\protect\citeauthoryear{Bowman, Gauthier, Rastogi, Gupta, Manning, and
  Potts}{Bowman et~al\mbox{.}}{2016}]%
        {bowman2016fast}
\bibfield{author}{\bibinfo{person}{Samuel~R Bowman}, \bibinfo{person}{Jon
  Gauthier}, \bibinfo{person}{Abhinav Rastogi}, \bibinfo{person}{Raghav Gupta},
  \bibinfo{person}{Christopher~D Manning}, {and} \bibinfo{person}{Christopher
  Potts}.} \bibinfo{year}{2016}\natexlab{}.
\newblock \showarticletitle{A fast unified model for parsing and sentence
  understanding}.
\newblock \bibinfo{journal}{\emph{arXiv preprint arXiv:1603.06021}}
  (\bibinfo{year}{2016}).
\newblock


\bibitem[\protect\citeauthoryear{Bustos and Pertusa}{Bustos and
  Pertusa}{2018}]%
        {bustos2018learning}
\bibfield{author}{\bibinfo{person}{Aurelia Bustos} {and}
  \bibinfo{person}{Antonio Pertusa}.} \bibinfo{year}{2018}\natexlab{}.
\newblock \showarticletitle{Learning eligibility in cancer clinical trials
  using deep neural networks}.
\newblock \bibinfo{journal}{\emph{Applied Sciences}} \bibinfo{volume}{8},
  \bibinfo{number}{7} (\bibinfo{year}{2018}), \bibinfo{pages}{1206}.
\newblock


\bibitem[\protect\citeauthoryear{Choi, Xiao, Stewart, and Sun}{Choi
  et~al\mbox{.}}{2018}]%
        {choi2018mime}
\bibfield{author}{\bibinfo{person}{Edward Choi}, \bibinfo{person}{Cao Xiao},
  \bibinfo{person}{Walter Stewart}, {and} \bibinfo{person}{Jimeng Sun}.}
  \bibinfo{year}{2018}\natexlab{}.
\newblock \showarticletitle{Mime: Multilevel medical embedding of electronic
  health records for predictive healthcare}. In
  \bibinfo{booktitle}{\emph{Advances in Neural Information Processing
  Systems}}. \bibinfo{pages}{4547--4557}.
\newblock


\bibitem[\protect\citeauthoryear{Conneau, Kiela, Schwenk, Barrault, and
  Bordes}{Conneau et~al\mbox{.}}{2017}]%
        {conneau2017supervised}
\bibfield{author}{\bibinfo{person}{Alexis Conneau}, \bibinfo{person}{Douwe
  Kiela}, \bibinfo{person}{Holger Schwenk}, \bibinfo{person}{Loic Barrault},
  {and} \bibinfo{person}{Antoine Bordes}.} \bibinfo{year}{2017}\natexlab{}.
\newblock \showarticletitle{Supervised learning of universal sentence
  representations from natural language inference data}.
\newblock \bibinfo{journal}{\emph{arXiv preprint arXiv:1705.02364}}
  (\bibinfo{year}{2017}).
\newblock


\bibitem[\protect\citeauthoryear{Dagan, Roth, Sammons, and Zanzotto}{Dagan
  et~al\mbox{.}}{2013}]%
        {dagan2013recognizing}
\bibfield{author}{\bibinfo{person}{Ido Dagan}, \bibinfo{person}{Dan Roth},
  \bibinfo{person}{Mark Sammons}, {and} \bibinfo{person}{Fabio~Massimo
  Zanzotto}.} \bibinfo{year}{2013}\natexlab{}.
\newblock \showarticletitle{Recognizing textual entailment: Models and
  applications}.
\newblock \bibinfo{journal}{\emph{Synthesis Lectures on Human Language
  Technologies}} \bibinfo{volume}{6}, \bibinfo{number}{4}
  (\bibinfo{year}{2013}), \bibinfo{pages}{1--220}.
\newblock


\bibitem[\protect\citeauthoryear{Devlin, Chang, Lee, and Toutanova}{Devlin
  et~al\mbox{.}}{2018}]%
        {devlin2018bert}
\bibfield{author}{\bibinfo{person}{Jacob Devlin}, \bibinfo{person}{Ming-Wei
  Chang}, \bibinfo{person}{Kenton Lee}, {and} \bibinfo{person}{Kristina
  Toutanova}.} \bibinfo{year}{2018}\natexlab{}.
\newblock \showarticletitle{Bert: Pre-training of deep bidirectional
  transformers for language understanding}.
\newblock \bibinfo{journal}{\emph{arXiv preprint arXiv:1810.04805}}
  (\bibinfo{year}{2018}).
\newblock


\bibitem[\protect\citeauthoryear{Duchi, Hazan, and Singer}{Duchi
  et~al\mbox{.}}{2011}]%
        {Duchi:2011:ASM:1953048.2021068}
\bibfield{author}{\bibinfo{person}{John Duchi}, \bibinfo{person}{Elad Hazan},
  {and} \bibinfo{person}{Yoram Singer}.} \bibinfo{year}{2011}\natexlab{}.
\newblock \showarticletitle{Adaptive Subgradient Methods for Online Learning
  and Stochastic Optimization}.
\newblock \bibinfo{journal}{\emph{J. Mach. Learn. Res.}}  \bibinfo{volume}{12}
  (\bibinfo{date}{July} \bibinfo{year}{2011}), \bibinfo{pages}{2121--2159}.
\newblock
\showISSN{1532-4435}
\urldef\tempurl%
\url{http://dl.acm.org/citation.cfm?id=1953048.2021068}
\showURL{%
\tempurl}


\bibitem[\protect\citeauthoryear{Ho, Ibrahim, Pal, Berberich, and Weikum}{Ho
  et~al\mbox{.}}{2019}]%
        {ho2019qsearch}
\bibfield{author}{\bibinfo{person}{Vinh~Thinh Ho}, \bibinfo{person}{Yusra
  Ibrahim}, \bibinfo{person}{Koninika Pal}, \bibinfo{person}{Klaus Berberich},
  {and} \bibinfo{person}{Gerhard Weikum}.} \bibinfo{year}{2019}\natexlab{}.
\newblock \showarticletitle{Qsearch: Answering Quantity Queries from Text}. In
  \bibinfo{booktitle}{\emph{International Semantic Web Conference}}. Springer,
  \bibinfo{pages}{237--257}.
\newblock


\bibitem[\protect\citeauthoryear{Hochreiter and Schmidhuber}{Hochreiter and
  Schmidhuber}{1997}]%
        {hochreiter1997long}
\bibfield{author}{\bibinfo{person}{Sepp Hochreiter} {and}
  \bibinfo{person}{J{\"u}rgen Schmidhuber}.} \bibinfo{year}{1997}\natexlab{}.
\newblock \showarticletitle{Long short-term memory}.
\newblock \bibinfo{journal}{\emph{Neural computation}} \bibinfo{volume}{9},
  \bibinfo{number}{8} (\bibinfo{year}{1997}), \bibinfo{pages}{1735--1780}.
\newblock


\bibitem[\protect\citeauthoryear{Hosmer~Jr, Lemeshow, and Sturdivant}{Hosmer~Jr
  et~al\mbox{.}}{2013}]%
        {hosmer2013applied}
\bibfield{author}{\bibinfo{person}{David~W Hosmer~Jr}, \bibinfo{person}{Stanley
  Lemeshow}, {and} \bibinfo{person}{Rodney~X Sturdivant}.}
  \bibinfo{year}{2013}\natexlab{}.
\newblock \bibinfo{booktitle}{\emph{Applied logistic regression}}.
  Vol.~\bibinfo{volume}{398}.
\newblock \bibinfo{publisher}{John Wiley \& Sons}.
\newblock


\bibitem[\protect\citeauthoryear{Huang, Wang, and Wang}{Huang
  et~al\mbox{.}}{2017}]%
        {huang2017instance}
\bibfield{author}{\bibinfo{person}{Yan Huang}, \bibinfo{person}{Wei Wang},
  {and} \bibinfo{person}{Liang Wang}.} \bibinfo{year}{2017}\natexlab{}.
\newblock \showarticletitle{Instance-aware image and sentence matching with
  selective multimodal lstm}. In \bibinfo{booktitle}{\emph{Proceedings of the
  IEEE Conference on Computer Vision and Pattern Recognition}}.
  \bibinfo{pages}{2310--2318}.
\newblock


\bibitem[\protect\citeauthoryear{Kang, Zhang, Tang, Hruby, Rusanov, Elhadad,
  and Weng}{Kang et~al\mbox{.}}{2017}]%
        {kang2017eliie}
\bibfield{author}{\bibinfo{person}{Tian Kang}, \bibinfo{person}{Shaodian
  Zhang}, \bibinfo{person}{Youlan Tang}, \bibinfo{person}{Gregory~W Hruby},
  \bibinfo{person}{Alexander Rusanov}, \bibinfo{person}{No{\'e}mie Elhadad},
  {and} \bibinfo{person}{Chunhua Weng}.} \bibinfo{year}{2017}\natexlab{}.
\newblock \showarticletitle{EliIE: An open-source information extraction system
  for clinical trial eligibility criteria}.
\newblock \bibinfo{journal}{\emph{Journal of the American Medical Informatics
  Association}} \bibinfo{volume}{24}, \bibinfo{number}{6}
  (\bibinfo{year}{2017}), \bibinfo{pages}{1062--1071}.
\newblock


\bibitem[\protect\citeauthoryear{Li, Xiao, Li, Yang, and Wang}{Li
  et~al\mbox{.}}{2017}]%
        {li2017identity}
\bibfield{author}{\bibinfo{person}{Shuang Li}, \bibinfo{person}{Tong Xiao},
  \bibinfo{person}{Hongsheng Li}, \bibinfo{person}{Wei Yang}, {and}
  \bibinfo{person}{Xiaogang Wang}.} \bibinfo{year}{2017}\natexlab{}.
\newblock \showarticletitle{Identity-aware textual-visual matching with latent
  co-attention}. In \bibinfo{booktitle}{\emph{Proceedings of the IEEE
  International Conference on Computer Vision}}. \bibinfo{pages}{1890--1899}.
\newblock


\bibitem[\protect\citeauthoryear{Meltzer and Noble}{Meltzer and Noble}{2008}]%
        {Meltzer2008}
\bibfield{author}{\bibinfo{person}{Eric Meltzer} {and} \bibinfo{person}{Paul
  Noble}.} \bibinfo{year}{2008}\natexlab{}.
\newblock \showarticletitle{Idiopathic pulmonary fibrosis}.
\newblock \bibinfo{journal}{\emph{Orphanet Journal of Rare Diseases}}
  (\bibinfo{year}{2008}).
\newblock


\bibitem[\protect\citeauthoryear{Mou, Men, Li, Xu, Zhang, Yan, and Jin}{Mou
  et~al\mbox{.}}{2015}]%
        {mou2015natural}
\bibfield{author}{\bibinfo{person}{Lili Mou}, \bibinfo{person}{Rui Men},
  \bibinfo{person}{Ge Li}, \bibinfo{person}{Yan Xu}, \bibinfo{person}{Lu
  Zhang}, \bibinfo{person}{Rui Yan}, {and} \bibinfo{person}{Zhi Jin}.}
  \bibinfo{year}{2015}\natexlab{}.
\newblock \showarticletitle{Natural language inference by tree-based
  convolution and heuristic matching}.
\newblock \bibinfo{journal}{\emph{arXiv preprint arXiv:1512.08422}}
  (\bibinfo{year}{2015}).
\newblock


\bibitem[\protect\citeauthoryear{Parikh, T{\"a}ckstr{\"o}m, Das, and
  Uszkoreit}{Parikh et~al\mbox{.}}{2016}]%
        {parikh2016decomposable}
\bibfield{author}{\bibinfo{person}{Ankur~P Parikh}, \bibinfo{person}{Oscar
  T{\"a}ckstr{\"o}m}, \bibinfo{person}{Dipanjan Das}, {and}
  \bibinfo{person}{Jakob Uszkoreit}.} \bibinfo{year}{2016}\natexlab{}.
\newblock \showarticletitle{A decomposable attention model for natural language
  inference}.
\newblock \bibinfo{journal}{\emph{arXiv preprint arXiv:1606.01933}}
  (\bibinfo{year}{2016}).
\newblock


\bibitem[\protect\citeauthoryear{Paszke, Gross, Chintala, Chanan, Yang, DeVito,
  Lin, Desmaison, Antiga, and Lerer}{Paszke et~al\mbox{.}}{2017}]%
        {paszke2017automatic}
\bibfield{author}{\bibinfo{person}{Adam Paszke}, \bibinfo{person}{Sam Gross},
  \bibinfo{person}{Soumith Chintala}, \bibinfo{person}{Gregory Chanan},
  \bibinfo{person}{Edward Yang}, \bibinfo{person}{Zachary DeVito},
  \bibinfo{person}{Zeming Lin}, \bibinfo{person}{Alban Desmaison},
  \bibinfo{person}{Luca Antiga}, {and} \bibinfo{person}{Adam Lerer}.}
  \bibinfo{year}{2017}\natexlab{}.
\newblock \showarticletitle{Automatic differentiation in PyTorch}.
\newblock  (\bibinfo{year}{2017}).
\newblock


\bibitem[\protect\citeauthoryear{Pennington, Socher, and Manning}{Pennington
  et~al\mbox{.}}{2014}]%
        {Pennington14glove:global}
\bibfield{author}{\bibinfo{person}{Jeffrey Pennington},
  \bibinfo{person}{Richard Socher}, {and} \bibinfo{person}{Christopher~D.
  Manning}.} \bibinfo{year}{2014}\natexlab{}.
\newblock \showarticletitle{Glove: Global vectors for word representation}. In
  \bibinfo{booktitle}{\emph{In EMNLP}}.
\newblock


\bibitem[\protect\citeauthoryear{Peters, Neumann, Iyyer, Gardner, Clark, Lee,
  and Zettlemoyer}{Peters et~al\mbox{.}}{2018}]%
        {peters2018deep}
\bibfield{author}{\bibinfo{person}{Matthew~E Peters}, \bibinfo{person}{Mark
  Neumann}, \bibinfo{person}{Mohit Iyyer}, \bibinfo{person}{Matt Gardner},
  \bibinfo{person}{Christopher Clark}, \bibinfo{person}{Kenton Lee}, {and}
  \bibinfo{person}{Luke Zettlemoyer}.} \bibinfo{year}{2018}\natexlab{}.
\newblock \showarticletitle{Deep contextualized word representations}.
\newblock \bibinfo{journal}{\emph{arXiv preprint arXiv:1802.05365}}
  (\bibinfo{year}{2018}).
\newblock


\bibitem[\protect\citeauthoryear{Radford, Narasimhan, Salimans, and
  Sutskever}{Radford et~al\mbox{.}}{2018}]%
        {radford2018improving}
\bibfield{author}{\bibinfo{person}{Alec Radford}, \bibinfo{person}{Karthik
  Narasimhan}, \bibinfo{person}{Tim Salimans}, {and} \bibinfo{person}{Ilya
  Sutskever}.} \bibinfo{year}{2018}\natexlab{}.
\newblock \showarticletitle{Improving language understanding by generative
  pre-training}.
\newblock  (\bibinfo{year}{2018}).
\newblock


\bibitem[\protect\citeauthoryear{Rockt{\"a}schel, Grefenstette, Hermann,
  Ko{\v{c}}isk{\`y}, and Blunsom}{Rockt{\"a}schel et~al\mbox{.}}{2015}]%
        {rocktaschel2015reasoning}
\bibfield{author}{\bibinfo{person}{Tim Rockt{\"a}schel},
  \bibinfo{person}{Edward Grefenstette}, \bibinfo{person}{Karl~Moritz Hermann},
  \bibinfo{person}{Tom{\'a}{\v{s}} Ko{\v{c}}isk{\`y}}, {and}
  \bibinfo{person}{Phil Blunsom}.} \bibinfo{year}{2015}\natexlab{}.
\newblock \showarticletitle{Reasoning about entailment with neural attention}.
\newblock \bibinfo{journal}{\emph{arXiv preprint arXiv:1509.06664}}
  (\bibinfo{year}{2015}).
\newblock


\bibitem[\protect\citeauthoryear{Roy, Vieira, and Roth}{Roy
  et~al\mbox{.}}{2015}]%
        {roy2015reasoning}
\bibfield{author}{\bibinfo{person}{Subhro Roy}, \bibinfo{person}{Tim Vieira},
  {and} \bibinfo{person}{Dan Roth}.} \bibinfo{year}{2015}\natexlab{}.
\newblock \showarticletitle{Reasoning about quantities in natural language}.
\newblock \bibinfo{journal}{\emph{Transactions of the Association for
  Computational Linguistics}}  \bibinfo{volume}{3} (\bibinfo{year}{2015}),
  \bibinfo{pages}{1--13}.
\newblock


\bibitem[\protect\citeauthoryear{Shang, Ma, Xiao, and Sun}{Shang
  et~al\mbox{.}}{2019}]%
        {ijcai2019-825}
\bibfield{author}{\bibinfo{person}{Junyuan Shang}, \bibinfo{person}{Tengfei
  Ma}, \bibinfo{person}{Cao Xiao}, {and} \bibinfo{person}{Jimeng Sun}.}
  \bibinfo{year}{2019}\natexlab{}.
\newblock \showarticletitle{Pre-training of Graph Augmented Transformers for
  Medication Recommendation}. In \bibinfo{booktitle}{\emph{Proceedings of the
  Twenty-Eighth International Joint Conference on Artificial Intelligence,
  {IJCAI-19}}}. \bibinfo{publisher}{International Joint Conferences on
  Artificial Intelligence Organization}, \bibinfo{pages}{5953--5959}.
\newblock
\urldef\tempurl%
\url{https://doi.org/10.24963/ijcai.2019/825}
\showDOI{\tempurl}


\bibitem[\protect\citeauthoryear{Smiley and Pugh}{Smiley and Pugh}{2011}]%
        {smiley2011apache}
\bibfield{author}{\bibinfo{person}{David Smiley} {and}
  \bibinfo{person}{David~Eric Pugh}.} \bibinfo{year}{2011}\natexlab{}.
\newblock \bibinfo{booktitle}{\emph{Apache Solr 3 Enterprise Search Server}}.
\newblock \bibinfo{publisher}{Packt Publishing Ltd}.
\newblock


\bibitem[\protect\citeauthoryear{Srivastava, Hinton, Krizhevsky, Sutskever, and
  Salakhutdinov}{Srivastava et~al\mbox{.}}{2014}]%
        {srivastava2014dropout}
\bibfield{author}{\bibinfo{person}{Nitish Srivastava},
  \bibinfo{person}{Geoffrey Hinton}, \bibinfo{person}{Alex Krizhevsky},
  \bibinfo{person}{Ilya Sutskever}, {and} \bibinfo{person}{Ruslan
  Salakhutdinov}.} \bibinfo{year}{2014}\natexlab{}.
\newblock \showarticletitle{Dropout: A simple way to prevent neural networks
  from overfitting}.
\newblock \bibinfo{journal}{\emph{The Journal of Machine Learning Research}}
  \bibinfo{volume}{15}, \bibinfo{number}{1} (\bibinfo{year}{2014}),
  \bibinfo{pages}{1929--1958}.
\newblock


\bibitem[\protect\citeauthoryear{Vaswani, Shazeer, Parmar, Uszkoreit, Jones,
  Gomez, Kaiser, and Polosukhin}{Vaswani et~al\mbox{.}}{2017}]%
        {vaswani2017attention}
\bibfield{author}{\bibinfo{person}{Ashish Vaswani}, \bibinfo{person}{Noam
  Shazeer}, \bibinfo{person}{Niki Parmar}, \bibinfo{person}{Jakob Uszkoreit},
  \bibinfo{person}{Llion Jones}, \bibinfo{person}{Aidan~N Gomez},
  \bibinfo{person}{{\L}ukasz Kaiser}, {and} \bibinfo{person}{Illia
  Polosukhin}.} \bibinfo{year}{2017}\natexlab{}.
\newblock \showarticletitle{Attention is all you need}. In
  \bibinfo{booktitle}{\emph{Advances in neural information processing
  systems}}. \bibinfo{pages}{5998--6008}.
\newblock


\bibitem[\protect\citeauthoryear{Wang, Li, Huang, and Lazebnik}{Wang
  et~al\mbox{.}}{2018}]%
        {wang2018learning}
\bibfield{author}{\bibinfo{person}{Liwei Wang}, \bibinfo{person}{Yin Li},
  \bibinfo{person}{Jing Huang}, {and} \bibinfo{person}{Svetlana Lazebnik}.}
  \bibinfo{year}{2018}\natexlab{}.
\newblock \showarticletitle{Learning two-branch neural networks for image-text
  matching tasks}.
\newblock \bibinfo{journal}{\emph{IEEE Transactions on Pattern Analysis and
  Machine Intelligence}} \bibinfo{volume}{41}, \bibinfo{number}{2}
  (\bibinfo{year}{2018}), \bibinfo{pages}{394--407}.
\newblock


\bibitem[\protect\citeauthoryear{Wang, Li, and Lazebnik}{Wang
  et~al\mbox{.}}{2016}]%
        {wang2016learning}
\bibfield{author}{\bibinfo{person}{Liwei Wang}, \bibinfo{person}{Yin Li}, {and}
  \bibinfo{person}{Svetlana Lazebnik}.} \bibinfo{year}{2016}\natexlab{}.
\newblock \showarticletitle{Learning deep structure-preserving image-text
  embeddings}. In \bibinfo{booktitle}{\emph{Proceedings of the IEEE conference
  on computer vision and pattern recognition}}. \bibinfo{pages}{5005--5013}.
\newblock


\bibitem[\protect\citeauthoryear{Wang and Jiang}{Wang and Jiang}{2015}]%
        {wang2015learning}
\bibfield{author}{\bibinfo{person}{Shuohang Wang} {and} \bibinfo{person}{Jing
  Jiang}.} \bibinfo{year}{2015}\natexlab{}.
\newblock \showarticletitle{Learning natural language inference with LSTM}.
\newblock \bibinfo{journal}{\emph{arXiv preprint arXiv:1512.08849}}
  (\bibinfo{year}{2015}).
\newblock


\bibitem[\protect\citeauthoryear{Weng, Tu, Sim, and Richesson}{Weng
  et~al\mbox{.}}{2010}]%
        {weng2010formal}
\bibfield{author}{\bibinfo{person}{Chunhua Weng}, \bibinfo{person}{Samson~W
  Tu}, \bibinfo{person}{Ida Sim}, {and} \bibinfo{person}{Rachel Richesson}.}
  \bibinfo{year}{2010}\natexlab{}.
\newblock \showarticletitle{Formal representation of eligibility criteria: a
  literature review}.
\newblock \bibinfo{journal}{\emph{Journal of biomedical informatics}}
  \bibinfo{volume}{43}, \bibinfo{number}{3} (\bibinfo{year}{2010}),
  \bibinfo{pages}{451--467}.
\newblock


\bibitem[\protect\citeauthoryear{Weng, Wu, Luo, Boland, Theodoratos, and
  Johnson}{Weng et~al\mbox{.}}{2011}]%
        {weng2011elixr}
\bibfield{author}{\bibinfo{person}{Chunhua Weng}, \bibinfo{person}{Xiaoying
  Wu}, \bibinfo{person}{Zhihui Luo}, \bibinfo{person}{Mary~Regina Boland},
  \bibinfo{person}{Dimitri Theodoratos}, {and} \bibinfo{person}{Stephen~B
  Johnson}.} \bibinfo{year}{2011}\natexlab{}.
\newblock \showarticletitle{EliXR: an approach to eligibility criteria
  extraction and representation}.
\newblock \bibinfo{journal}{\emph{Journal of the American Medical Informatics
  Association}} \bibinfo{volume}{18}, \bibinfo{number}{Supplement\_1}
  (\bibinfo{year}{2011}), \bibinfo{pages}{i116--i124}.
\newblock


\bibitem[\protect\citeauthoryear{Yuan, Ryan, Ta, Guo, Li, Hardin, Makadia, Jin,
  Shang, Kang, et~al\mbox{.}}{Yuan et~al\mbox{.}}{2019}]%
        {yuan2019criteria2query}
\bibfield{author}{\bibinfo{person}{Chi Yuan}, \bibinfo{person}{Patrick~B Ryan},
  \bibinfo{person}{Casey Ta}, \bibinfo{person}{Yixuan Guo},
  \bibinfo{person}{Ziran Li}, \bibinfo{person}{Jill Hardin},
  \bibinfo{person}{Rupa Makadia}, \bibinfo{person}{Peng Jin},
  \bibinfo{person}{Ning Shang}, \bibinfo{person}{Tian Kang}, {et~al\mbox{.}}}
  \bibinfo{year}{2019}\natexlab{}.
\newblock \showarticletitle{Criteria2Query: a natural language interface to
  clinical databases for cohort definition}.
\newblock \bibinfo{journal}{\emph{Journal of the American Medical Informatics
  Association}} \bibinfo{volume}{26}, \bibinfo{number}{4}
  (\bibinfo{year}{2019}), \bibinfo{pages}{294--305}.
\newblock


\end{thebibliography}

\end{document}